\documentclass[letterpaper]{article} 
\usepackage{aaai24}  
\usepackage{times}  
\usepackage{helvet}  
\usepackage{courier}  
\usepackage[hyphens]{url}  
\usepackage{graphicx} 
\usepackage{natbib}  
\usepackage{caption} 
\usepackage{algorithm}
\usepackage{algorithmic}

\usepackage{newfloat}
\usepackage{listings}
\DeclareCaptionStyle{ruled}{labelfont=normalfont,labelsep=colon,strut=off} 
\floatstyle{ruled}
\newfloat{listing}{tb}{lst}{}
\floatname{listing}{Listing}
\title{Supervision Interpolation via LossMix:\\Generalizing Mixup for Object Detection and Beyond}
\author{
    Thanh Vu\textsuperscript{\rm 1,2}\thanks{Work done during a residency with Mineral.},
    Baochen Sun\textsuperscript{\rm 2}\thanks{Project lead.},
    Bodi Yuan\textsuperscript{\rm 2},
    Alex Ngai\textsuperscript{\rm 2}
    Yueqi Li\textsuperscript{\rm 2
    },
    Jan-Michael Frahm\textsuperscript{\rm 1}
}
\affiliations{
    \textsuperscript{\rm 1}University of North Carolina at Chapel Hill\\
    \textsuperscript{\rm 2}Mineral\\
    \{thanhvu,baochens,bodiyuan\}@mineral.ai,
    \{alexander.s.ngai,yueqili.innovation\}@gmail.com,
    jmf@cs.unc.edu
}

\usepackage{amsmath}
\usepackage{enumitem} 
\usepackage{overpic} 
\usepackage{color}
\usepackage{multirow}
\usepackage{pifont}

\newcommand{\Fig}[1]{Fig.~\ref{fig:#1}}
\newcommand{\Figure}[1]{Figure~\ref{fig:#1}}
\newcommand{\Tab}[1]{Tab.~\ref{tab:#1}}
\newcommand{\Table}[1]{Table~\ref{tab:#1}}

\newcommand{\Eq}[1]{Eq.~\ref{eq:#1}}
\newcommand{\Equation}[1]{Equation~\ref{eq:#1}}
\newcommand{\Sec}[1]{Sec.~\ref{sec:#1}}

\newcommand{\cmark}{\ding{51}}%
\newcommand{\xmark}{\ding{55}}%

\usepackage{booktabs}

\begin{document}

\maketitle

\begin{abstract}
The success of data mixing augmentations in image classification tasks has been well-received. However, these techniques cannot be readily applied to object detection due to challenges such as spatial misalignment, foreground/background distinction, and plurality of instances. To tackle these issues, we first introduce a novel conceptual framework called Supervision Interpolation (SI), which offers a fresh perspective on interpolation-based augmentations by relaxing and generalizing Mixup. Based on SI, we propose LossMix, a simple yet versatile and effective regularization that enhances the performance and robustness of object detectors and more. Our key insight is that we can effectively regularize the training on mixed data by interpolating their loss errors instead of ground truth labels. Empirical results on the PASCAL VOC and MS COCO datasets demonstrate that LossMix can consistently outperform state-of-the-art methods widely adopted for detection. Furthermore, by jointly leveraging LossMix with unsupervised domain adaptation, we successfully improve existing approaches and set a new state of the art for cross-domain object detection.
\end{abstract}


\section{{Introduction}}
\label{sec:intro}
Over the past decade, object detection has made remarkable progress, with impressive scores on challenging benchmarks such as MS COCO~\cite{coco}. 
However, state-of-the-art detectors still suffer from poor generalization abilities and struggle with data outside their training distribution, especially under domain shifts~\cite{survey_da_det_ssci_20,survey_uda_det_21}. 
Recently, data mixing techniques, pioneered by Mixup~\cite{mixup}, have emerged as an effective augmentation and regularization method for improving accuracy and robustness in deep neural networks.
These techniques~\citep{
mixup,
cutmix,
manifold_mixup,
puzzlemix,
supermix,
stylemix} 
use a linear interpolation of both images and their labels to generate synthetic training data.
The ``mixing" process encourages the model to behave linearly between training examples, which can potentially reduce undesired oscillations for out-of-distribution predictions.
Since its introduction in 2018, Mixup has garnered increasing attention and has been widely adopted for image classification problems~\citep{
mixup,
cutmix,
manifold_mixup,
mixup_cls_adv,
puzzlemix,
mixup_cls_dual,
supermix,
stylemix,
mixup_cls_fixbi,
tokenmix,
automix,
regmixup,
over_train_mixup}.
This motivates us to investigate Mixup-like augmentation for object detection. 

\begin{figure}  
    \begin{center}
    \includegraphics[width=\linewidth]{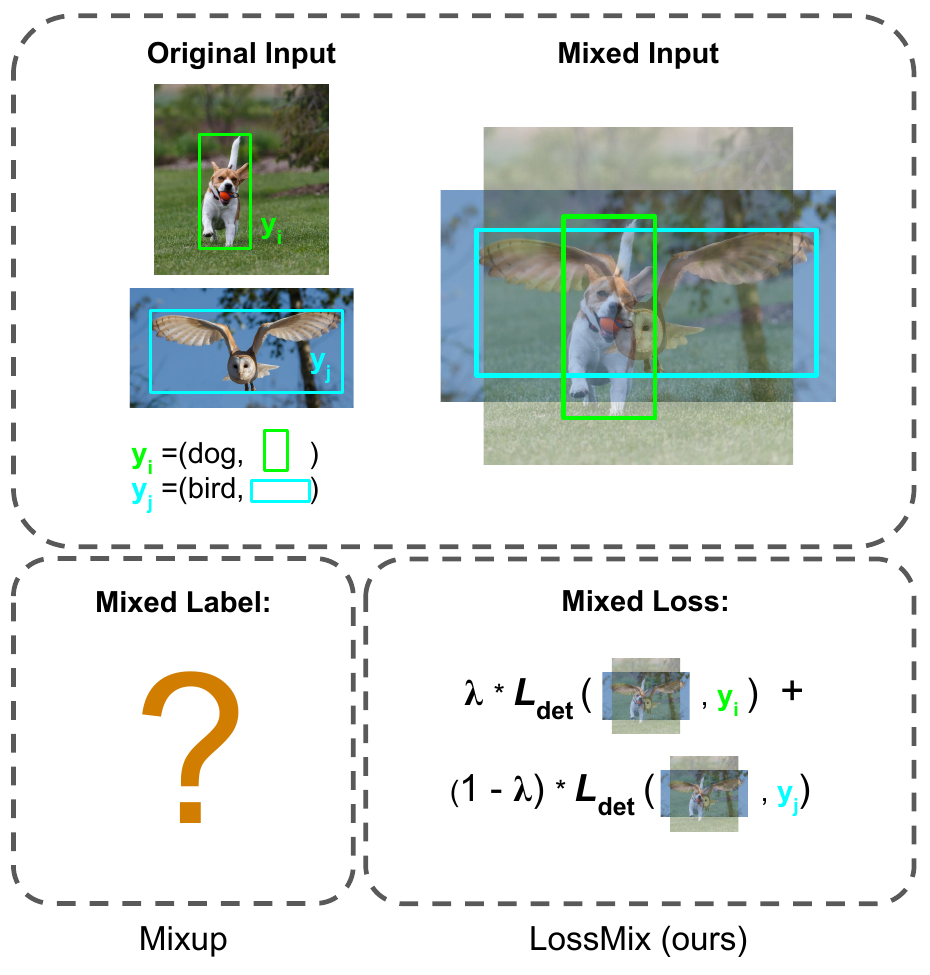}
    \end{center}
    \caption{
        Compared to label-mixing (e.g., Mixup, CutMix), the proposed LossMix deploys interpolated losses instead of interpolated ground truths as the mixed supervision signals. This significantly simplifies the challenges involved in applying data mixing to higher-level tasks such as detection.
    }
    \label{fig:teaser}
\end{figure}

\begin{figure}
    \begin{center}
    \includegraphics[width=\linewidth]{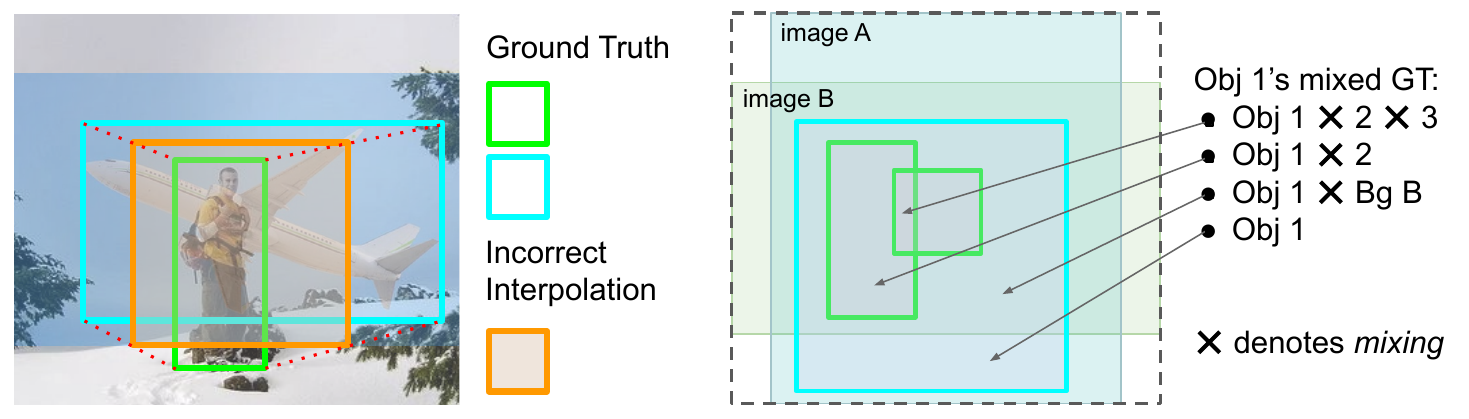}
    \end{center}
    \caption{
        Challenges of applying Mixup to object detection. 
        Left: Semantically incorrect interpolation of bounding box coordinates.
        Right: Complication of mixed category labels.
    }
    \label{fig:wrong_bbox_mixing}
\end{figure}

Unfortunately, Mixup cannot be applied to object detection task off the shelf.
On one hand, the mixing of the category label of object instances is non-trivial (\Fig{wrong_bbox_mixing}) due to issues such as spatial misalignment, foreground/background distinctions, and the plurality of instances.
In contrast to classification, where images share the same shape and each only has one class label, object detectors need to handle images and objects of different aspect ratios and positions. 
This makes it impossible to guarantee the alignment of mixed objects and creates much more complexity for the interpolation of class labels.
\Fig{wrong_bbox_mixing} (left) provides a visualization of these challenges.
In addition, ground truth object detection annotations are composed of bounding box coordinates that cannot be naively interpolated without disturbing the localization ground truth (\Fig{wrong_bbox_mixing}, right).

\begin{figure}
    \begin{center}
    \includegraphics[width=\linewidth]{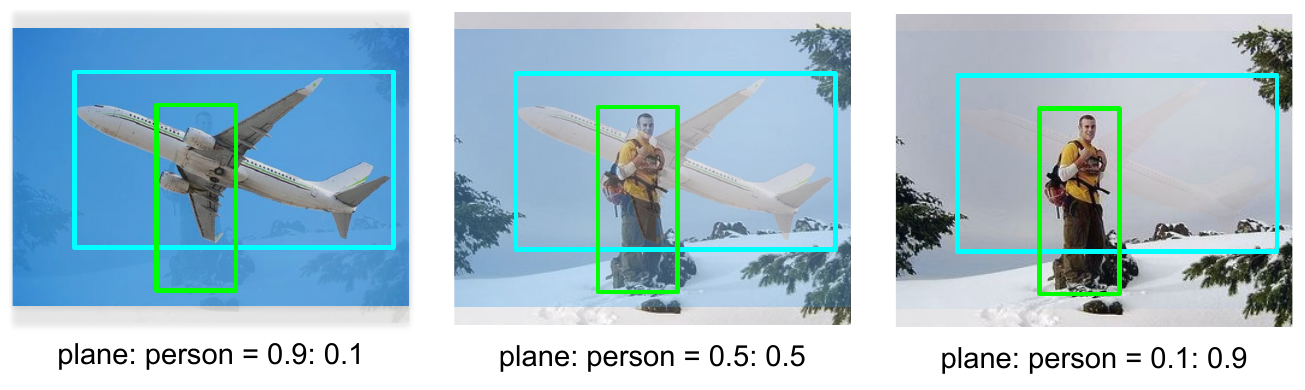}
    \end{center}
    \caption{
        The state-of-the-art \textit{Union} strategy is problematic because all ground truth bounding boxes are treated equally regardless of the mixing ratio.
    }
    \label{fig:union_issue}
\end{figure}

The current state-of-the-art approach~\citep{
bof,
yolov5,
yolox,
union_ssl_instant_teaching,
union_instseg_seesaw,
union_mot_bytetrack,
union_da_fewshot_acrofod,
union_ssl_dual,
small_obj_2023,
unmanned_2023,
yolov8} works around this by taking an unweighted, uniform \textit{union} of all bounding boxes as new ground truth for the augmented image.
Although this strategy has shown some success, there are several limitations. 
First, the approach does not follow the input-target dual interpolation principle that fuels the success of Mixup in classification, as it considers all component bounding boxes equally regardless of the actual mixing ratio $\lambda$.
Second, when small mixing coefficients are used, e.g. $\lambda < 0.1$, the \textit{Union} strategy can produce noisy mixed object labels (\Fig{union_issue}), potentially leading to sensitivity to noise and hallucinations in the model.
These approaches expect the models to be able to predict all object instances with equal likelihood, regardless of their visibility.
Finally, most of the previous studies have focused on using data mixing for semi-supervised~\cite{union_ssl_instant_teaching,union_ssl_dual} or few-shot learning~\cite{union_da_fewshot_acrofod}, rather than general object detection, aside from \cite{bof}.
More efforts exploring data mixing for general object detection are still needed to address these limitations.

To address these problems, we introduce two novel ideas: Supervision Interpolation (SI) and LossMix.
SI generalizes Mixup's input-target formulation by relaxing the requirement to explicitly interpolate the labels. 
Instead, we hypothesize that it is possible to interpolate other forms of target supervision besides explicitly augmenting the ground truth.
Based on this, we then propose LossMix, a simple but effective and versatile regularization that enables data mixing augmentation to strengthen object detection models and more.
Our key insight is that we can effectively interpolate the losses, instead of the ground truth labels, according to the input's interpolation.
Intuitively, from a data mixing perspective, LossMix interpolates the gradient signals that guide the models' learning, instead of explicitly augmenting the supervision labels like prior approaches~\cite{mixup,cutmix}.
From an object detection perspective, LossMix weights the penalty for each prediction based on their augmented visibility.
For example, LossMix would scale down the penalty for a failure to detect the plane with $\lambda=0.1$ in \Fig{union_issue} (right) since it has low visibility, while Union would treat both the plane and the person ($\lambda=0.9$) equally.
LossMix is flexible and can implicitly handle the mixing of sub-tasks (both classification and localization), while remaining true to the input-target dual interpolation principle that powered the success of the original Mixup~\cite{mixup}.
In short, our contributions are:
\begin{itemize}
    \item We introduce Supervision Interpolation (SI), a conceptual reinterpretation and generalization of Mixup~\cite{mixup}-like input-label interpolation formulation.
    \item Based on SI, we propose LossMix, a simple but effective and versatile regularization that enables direct data mixing augmentation for object detection.
    \item We demonstrate that LossMix consistently outperforms state-of-the-art mixing methods for object detection on PASCAL VOC~\shortcite{pascal} and MS COCO~\shortcite{coco} datasets. 
    \item We leverage LossMix to enhance the recent Adaptive Teacher~\cite{amt} framework and achieve a new state of the art for unsupervised domain adaptation.
\end{itemize}

\section{Related Work}
\label{sec:related}
\paragraph{Data Mixing Augmentations} 
The original Mixup~\cite{mixup} was designed for image classification tasks, proposing to use a convex combination of data and labels to expand the space of augmented training examples.
Mixup, CutMix~\cite{cutmix}, and follow-up works~\citep{
manifold_mixup,
mixup_cls_adv,
puzzlemix,
mixup_cls_dual,
supermix,
stylemix,
mixup_cls_fixbi,
lossmixup,
tokenmix,
automix,
regmixup,
alignmixup,
openmixup,
over_train_mixup}
have demonstrated the benefits of this interpolation-based augmentation for improving models' memorization and sensitivity to appearance changes. 
Since then, many works have utilized Mixup for image classification problems~\cite{mixup_cls_adv,mixup_cls_dual,mixup_cls_fixbi}.
However, due to the challenges discussed above, only a few studies~\cite{bof, yolov8} have explored the use of Mixup for general object detection.
Others tend to focus on settings with limited labeled data, such as semi-supervised learning~\cite{union_ssl_instant_teaching,union_ssl_dual}, few-shot learning~\cite{union_da_fewshot_acrofod}, domain adaptation~\cite{afan,union_da_fewshot_acrofod}, or specific applications~\cite{
unmanned_2023,
small_obj_2023}.

\paragraph{Cross-Domain Object Detection}
There are two main approaches for object detection: one-stage object detectors that attempt to perform localization and classification simultaneously~\cite{ssd,yolo}, and two-stage object detectors that first generate object proposals and then perform classification and bounding box refinement in the second stage~\cite{faster_rcnn}.
Domain Adaptation based approaches aims to build a robust detectors that can generalize well to a target domain with limited or no labeled data.
These methods can either explicitly align the feature distributions using a specific distance metric~\cite{long2015icml,long2016nips,sun2016aaai}, or implicitly align the distributions using an adversarial loss ~\cite{ganin2016jmlr,hoffman2018icml,long2018nips} or GAN~\cite{murez2018cvpr,lee2021cvpr}.
While most current works in domain adaptation focus on image classification~\cite{
da_det_image_stats,
da_cls_seg_pareto,
da_seg_style_fog,
da_cls_slimmable,
da_adamatch,
da_bridge_domain,
da_transformer_improve,
da_cls_dine,
da_crowd_cnt,
da_geotexaug,
da_cls_det_smooth,
da_transformer_safe}, a few have delved into object detection~\cite{
da_det_pit,
da_det_progressive,
umt,
da_det_ssal,
simrod,
amt}.
Data mixing is appealing in the context of UDA because of the opportunity to strategically blend cross-domain information during training.
To our best knowledge, the topic of data mixing for cross-domain object detection remains largely understudied~\cite{afan}.
In this work, we explore the application of LossMix to UDA.

\section{Methodology}

\paragraph{What does Mixup do?}
Mixup \cite{mixup} trains models with virtual examples constructed by convex combinations of pairs of examples and their associated labels:
\setlength{\abovedisplayskip}{6pt}
\setlength{\belowdisplayskip}{6pt}
\begin{equation} \label{eq:mixup_x}
    \tilde{x} = \lambda x_i + (1 - \lambda)x_j
\end{equation}
\begin{equation} \label{eq:mixup_y}
    \tilde{y} = \lambda y_i + (1 - \lambda)y_j 
\end{equation}
where $(x_i, y_i)$ and $(x_j , y_j)$ denote randomly sampled pairs of image and ground truth label and $\lambda \in [0, 1]$ denotes the interpolation coefficient.
Training with Mixup-augmented data entails minimizing the empirical vicinal risk:
\begin{equation} \label{eq:mixup_risk}
    \mathcal{R}_v(f) = \frac{1}{m} \sum_{i=1}^{m} \mathcal{L} (f(\tilde{x}), \tilde{y}) 
\end{equation}
The idea is to regularize the learning using a prior knowledge that linear interpolations of the input features (input mixing) should yield linear interpolations of the corresponding output (label mixing).
Such linear behavior in-between training examples can potentially mitigate undesired oscillations when predicting outside the training distribution.

\paragraph{Limitation of Label Mixing}
As discussed in the \Sec{intro} and illustrated in \Fig{wrong_bbox_mixing}, despite working well for classification, the problem quickly arise when considering Mixup for other higher-level tasks like object detection.
This is precisely because \Eq{mixup_y} imposes a hard constraint for the augmented labels $\tilde{y}$ to be an explicit linear combination of the real labels $y_i$ and $y_j$.
In object detection, for every $x_i$, the label $y_i$ contains a set of object annotations, each has their own class and bounding box coordinates: $y_i=\{(c_{i1},b_{i1}), (c_{i2},b_{i2}),\ldots\}$, with $(c_{ik},b_{ik})$ denoting an object instance with class label $c_{ik}$ and box coordinates $b_{ik}$.
This results in an ill-defined $\tilde{y}$ and makes label mixing exponentially more complicated.
Existing work~\cite{
bof,
union_instseg_seesaw,
union_ssl_instant_teaching,
union_ssl_dual,
union_da_fewshot_acrofod,
union_mot_bytetrack} chose to take an unweighted union of $y_i$ and $y_j$, yielding $\tilde{y}=\{(c_{i1},b_{i1}), (c_{i2},b_{i2}),\ldots,(c_{j1},b_{j1}), (c_{j2},b_{j2}),\ldots\}$.
Although this heuristic may offer some improvement in practice, it does not faithfully interpolate labels since $\tilde{y}$ is independent of $\lambda$ and may lead to sub-optimal results.
For example, small $\lambda$ could be problematic since some objects become barely visible (\Fig{union_issue}), creating noisy labels.

\subsection{Supervision Interpolation}
We identify that the root cause of the aforementioned issues for both Mixup and unweighted union strategy is the label mixing requirement defined in \Eq{mixup_y}.
Despite working well for simple classification tasks, this policy clearly creates much complications for higher-level tasks such as object detection.
To address this, we propose Supervision Interpolation (SI), a conceptual reinterpretation and generalization of Mixup's input-label interpolation formulation.
In SI, we train models using a dual of interpolated data $\tilde{x}$ and proportionally interpolated supervision signals.
Formally, SI trains models using convex combinations of examples and correspondingly augmented supervision signals: 
\begin{equation} \label{eq:sit_x}
    \tilde{x} = \lambda x_i + (1 - \lambda)x_j
\end{equation}
\begin{equation} \label{eq:sit_s}
    \tilde{S} = \lambda S_i + (1 - \lambda)S_j
\end{equation}
where $x$ denotes the training input, e.g., images, $S$ denotes the supervision signal, and $\lambda \in [0, 1]$ denotes the mixing coefficient.
Intuitively, SI regulates the training by interpolating the supervision or gradient signals guiding the model, depending on the interpolated inputs.
For example, Mixup-based augmentations~\cite{
mixup,
cutmix,
supermix,
stylemix,
automix} and ICT~\cite{interpolation_ssl_det,ict} in semi-supervised learning can be seen as a special case of SI where the supervision signals are the ground truth classification labels.
Based on such a flexible SI framework, next we will introduce, LossMix, a versatile method that enables data mixing and helps strengthen object detectors.

\subsection{LossMix}
Given the conceptual framework of Supervision Interpolation (SI), we then introduce LossMix, an equally simple but more versatile, task-agnostic sibling of Mixup that interpolates the loss errors instead of target labels.
Specifically, the LossMix-augmented data:
\begin{equation} \label{eq:our_x}
    \tilde{x} = \lambda x_i + (1 - \lambda)x_j
\end{equation}
\begin{equation} \label{eq:our_y}
    \tilde{y} = \{(y_i; \lambda), (y_j; (1 - \lambda)\}
\end{equation}
are coupled with an augmented loss function:
\begin{equation} \label{eq:our_loss}
    \tilde{\mathcal{L}} (f(\tilde{x}), \tilde{y}) = \lambda \mathcal{L} (f(\tilde{x}), y_i) 
     + (1 - \lambda) \mathcal{L} (f(\tilde{x}), y_j)
\end{equation}
From a SI perspective, our supervision signal is the loss errors weighted relative to $y_i$ and $y_j$, instead of an explicit interpolation of $y_i$ and $y_j$.
From the Mixup perspective, we have relaxed the constraint in \Eq{mixup_y}
and only characterise the virtual target $\tilde{y}$ using weighted $y_i$ and $y_j$ without explicitly constraint the form of $\tilde{y}$.
Note that as $\lambda \rightarrow 0.0$ or $1.0$, LossMix optimization will approach the standard empirical risk minimization.
We would like to highlight that LossMix is not only simple and effective, but also highly versatile:

\paragraph{Simple-yet-effective} 
The idea of loss mixing is straightforward and intuitive, both conceptually and implementation-wise, allowing easy adaptation to existing frameworks. Nonetheless, by design, LossMix helps circumvent the semantic collapse of bounding box interpolation (\Fig{wrong_bbox_mixing}) and approximation issues of unweighted union approach (\Fig{union_issue}).
Moreover, our experiments demonstrate the effectiveness and robustness of LossMix despite the simplicity in the design, successfully yielding improvement across two standard object detection datasets: PASCAL VOC~\cite{pascal} and MS COCO~\cite{coco}.
Finally, by applying LossMix to domain adaptation, we can further enhance state-of-the-art methods in cross-domain object detection~\cite{amt}.

\paragraph{Generalizability} LossMix leverages a versatile loss weighting formulation that is potentially applicable to different tasks (i.e., classification, detection, etc.) as well as different input mixing strategies (i.e., Mixup~\cite{mixup}, CutMix~\cite{mixup}, etc.).     
LossMix is loss-agnostic, which simplifies its application in object detection and can inspire more applications beyond cross- entropy loss based classification, e.g. regression tasks like localization or depth estimation.
The core idea of LossMix lies in the mixing of target loss signal $y$ and does not limit the input mixing. 
This opens the door for different strategies, 
including pixel-based~\cite{mixup}, region-based~\cite{cutmix}, style-based~\cite{stylemix} and more.

\subsection{LossMix in Action}
\paragraph{Classification \& Segmentation}
For standard classification and segmentation with Cross-Entropy loss, 
we can show that LossMix optimization is equivalent to that of Mixup:
\begin{align} \label{eq:ce_loss}
    & \tilde{\mathcal{L}}_{lossmix} (f(\tilde{x}), \tilde{y}_{lossmix}) 
    \\
    &= \lambda \mathcal{L}_{ce} (f(\tilde{x}), y_i)
     + (1-\lambda) \mathcal{L}_{ce} (f(\tilde{x}),y_j)
     \\
    &=  - \lambda y_i \log f(\tilde{x}) - (1-\lambda) y_j \log f(\tilde{x}) 
    \\
    &= - (\lambda y_i + (1-\lambda) y_j) \log f(\tilde{x}) 
    \\
    &= \mathcal{L}_{ce}(f(\tilde{x}), \lambda y_i + (1-\lambda) y_j)
    \\
    &= \tilde{\mathcal{L}}_{mixup} (f(\tilde{x}), \tilde{y}_{mixup})
\end{align}
This means LossMix enjoys the same advantages as Mixup when applied to CE-based classification and segmentation problems~\citep{
mixup,
cutmix,
mixup_cls_adv,
manifold_mixup,
mixup_cls_dual,
puzzlemix,
stylemix,
supermix,
mixup_cls_fixbi,
regmixup,
tokenmix,
automix,
openmixup,
over_train_mixup}.
It is worth noting that loss-mixing formulation for Mixup by itself is not new~\cite{mixup_code,automix}. 
What sets LossMix apart is its use of the Supervision Interpolation concept to replace label mixing with loss mixing at a fundamental level. 
This significantly enhances its generalizability, making CE-based classification a special case rather than the only option.
Since the benefits of LossMix/Mixup for classification are well studied, we focus on exploring LossMix for object detection and domain adaptation.

\paragraph{Object Detection}
Since LossMix makes no assumption about the loss functions $\mathcal{L}$ in \Eq{our_loss}, we can easily apply it to object detection by interpolating both the classification loss and box regression loss.
For example, for Faster RCNN~\cite{faster_rcnn}, $\mathcal{L}$ takes the form of the standard supervised loss for two-stage detectors:
\begin{align} \label{eq:loss_det}
\begin{split}
    \mathcal{L}_{det}(f(x),y) 
    &= \mathcal{L}^{rpn}(f(x),y) + \mathcal{L}^{roi}(f(x),y) \\
    &= \mathcal{L}_{cls}^{rpn}(f(x),y) + \mathcal{L}_{reg}^{rpn}(f(x),y) \\
    &+ \mathcal{L}_{cls}^{roi}(f(x),y) + \mathcal{L}_{reg}^{roi}(f(x),y)
\end{split}
\end{align}
Here, $\mathcal{L}^{rpn}$ denotes the loss of Region Proposal Network (RPN) which generates candidate proposals, while $\mathcal{L}^{roi}$ denotes the loss for Region of Interest (ROI) branch. Both branches perform bounding box regression and classification tasks, specifically binary classification for RPN (object or not) and multi-class classification for ROI \cite{faster_rcnn}.
Given a mixing coefficient $\lambda$, we can directly re-weight all sub-task losses as follows (omitting $f$ for brevity): 
\begin{equation} \label{eq:det_loss_mix}
    \tilde{\mathcal{L}}_{det} (\tilde{x}, \tilde{y}) = \lambda \mathcal{L}_{det} (\tilde{x}, y_i)
     + (1 - \lambda) \mathcal{L}_{det} (\tilde{x}, y_j)
\end{equation}

\paragraph{Domain Adaptation}
We apply LossMix alongside  the recent Adaptive Teacher~\cite{amt} (AT), a two-stage self-distillation method for cross-domain object detection.
During the \textit{Warmup} phase, we initialize both Teacher and Student models, who weights are shared, using standard object detection training with labeled source-domain data.
We leverage LossMix to mix intra-source domain data to encourage better (non-directional) generalization and improve robustness on unseen data. 
However, initializing with pure source domain data risks biasing the Teacher model towards such a distribution \cite{umt}, potentially yielding low-quality pseudo labels.
Thus, we use unlabeled target images to mitigate this, specifically by mixing a small amount (e.g. $\lambda < 0.1$) of them into the labeled source images.

During the \textit{Adaptation} phase, both models are jointly trained using the same cross-domain distillation~\cite{amt,umt}.
Thanks to the pseudo labels generated by the Teacher, we can perform intra-domain mixing with both labeled source (source-source) and pseudo-labeled target data (target-target).
Moreover, we also deploy a balanced version of the inter-domain mixing used during the warmup phase. 
The reasons are twofold.
First of all, with the presence of target pseudo labels, we now have the option to perform inter-domain mixing in the same manner as we do for intra-domain labeled source mixing, which is something not feasible during the warmup phase.
Secondly, the reason we do not want to continue using noise mixing is because it would become merely additional source domain data (the mixed-in unlabeled target image is only noise) and can potentially bias the model towards source distribution. 
Compare to this, the pseudo labels are much stronger signal that will push the model to learn target features.

\section{Experiments: Object Detection}
\subsection{Experimental Settings}
\paragraph{Datasets} 
We conduct experiments on two standard benchmark datasets in object detection, namely PASCAL VOC~\cite{pascal} and MS COCO~\cite{coco}.
We follow \cite{bof} and use the combination of PASCAL VOC 2007 \textit{trainval} (5k images) and 2012 \textit{trainval} (12k images) for training.
Together they make up 16,551 images of 20 categories of common, real-world objects, each with fully annotated bounding boxes and class labels.
The evaluation is done on PASCAL VOC 2007 \textit{test} set (5K images).
MS COCO~\cite{coco} is composed of  80 object categories and is 10 times larger than PASCAL VOC.
We train on \textit{train2017} (118K images) and evaluated on \textit{val2017} (5K images).

\paragraph{{Baseline models}}
We use three main baseline models to evaluate the performance of our proposed LossMix. 
The first one is a baseline, bare bone model without any Mixup-like data augmentation. 
Second, we compare LossMix against 
\textit{Union} mixing,
the current state-of-the-art approach widely used by prior studies~\citep{
bof,
yolov5,
yolox,
union_ssl_instant_teaching,
union_instseg_seesaw,
union_mot_bytetrack,
union_da_fewshot_acrofod,
union_ssl_dual,
small_obj_2023,
unmanned_2023,
yolov8} works around this by taking an unweighted, uniform \textit{union} of all bounding boxes as new ground truth for the augmented image. 
Finally, we also compare with the ``Noise" mixing strategy used by \cite{afan} for unsuperivsed domain adaptation. In a nutshell, it mixes input image A with a small amount of image B (e.g. $\lambda < 0.1$) acting only as color augmentation and discards any objects exists in B.

\paragraph{{Implementation Details}} 
We leverage the open-source PyTorch-based Detectron2~\cite{detectron2} repository as our object detection codebase for experimentation.
We use Faster RCNN~\cite{faster_rcnn} with ResNet~\cite{resnet}--FPN~\cite{fpn} backbone as our baseline model.
By default, ImageNet1K~\cite{imagenet} pretrained weights are used to initialize the networks.
Unless otherwise specified, we use a batch size of 64 for faster convergence, an initial learning rate of 0.08, and the default step scheduler from Detectron2.
We train PASCAL VOC for 18K  iterations, which is about 70 epochs, and MS COCO for 270K iterations, or roughly 146.4 epochs.
All experiments were trained with 8 NVIDIA GPUs, either V100 or A100.

\begingroup
\begin{table}[t]
    \centering
    \begin{tabular}{@{}ll|ccc@{}}
    \toprule
    Backbone
        & Method
        & AP
        & AP$_{50}$
        & AP$_{75}$\\
    \midrule\midrule
    ~
        & Baseline
        & 53.30
        & 78.89
        & 59.11\\
    ResNet-50
        & Noise
        & 54.36
        & 80.46
        & 59.92\\
    + FPN
        & Union
        & 55.05
        & 82.02
        & 61.72\\
    ~
        & LossMix (ours)
        & \textbf{55.87}
        & \textbf{82.44}
        & \textbf{62.88}\\
    \midrule
    ~
        & Baseline
        & 53.25
        & 79.87
        & 59.24\\
    ResNet-101
        & Noise
        & 54.90
        & 81.52
        & 60.78\\
    + FPN
        & Union
        & 55.01
        & 82.56
        & 61.50\\
    ~
        & LossMix (ours)
        & \textbf{55.91}
        & \textbf{82.84}
        & \textbf{62.72}\\
    \bottomrule
    \end{tabular}
    \caption{
        PASCAL VOC results with Faster RCNN detector and ResNet-50/101 FPN backbone.
        For each method, we report the best checkpoint based on AP50 metric following PASCAL VOC standard.
        Best results are in \textbf{bold}. 
        Our proposed LossMix outperforms state-of-the-art approaches such as Union and Noise to achieve the best overall results.
    } 
    \label{tab:det_voc}
\end{table}

\subsection{Results}
\paragraph{PASCAL VOC dataset}
\Tab{det_voc} shows the results for LossMix in comparison with the baseline model and prior methods on PASCAL VOC dataset.
First, we can see that all data mixing methods offer some improvements over the base Faster RCNN model, even ``Noise" despite the weak mixing augmentation.
This validates our interest in studying data mixing regularization for object detection.
Second, among the detectors that deploys different mixing strategies, those with LossMix clearly outperform others. 
Specifically, our method yields up to  $+0.9$AP compared to Union, $+1.5$AP compared to Noise, and $+2.7$AP compared to no-mixing baseline.
Overall, LossMix achieves the best performance across all three evaluation metrics, AP, AP$_{50}$, and AP$_{75}$, as well as both backbones, ResNet-50 and ResNet-101 FPN.

\begingroup
\begin{table*}[t]
    \centering
    \begin{tabular}{@{}ll|cccccc@{}}
    \toprule
    Backbone
        & Method
        & AP
        & AP$_{50}$
        & AP$_{75}$
        & AP$_{S}$
        & AP$_{M}$
        & AP$_{L}$\\ 
    \midrule\midrule
    \multirow{4}{*}{ResNet-50 + FPN}
        & Baseline
        & 40.41
        & 60.95
        & 43.95
        & 24.59
        & 43.82
        & 51.82\\
    
        & Noise
        & 41.01
        & 61.74
        & 44.98
        & 25.20
        & 44.71
        & 52.34\\
    
        & Union
        & 41.43
        & {\textbf{62.75}}
        & 45.68
        & {\textbf{25.59}}
        & 45.19
        & 52.61\\
    
        & LossMix (ours)
        & {\textbf{41.82}}
        & 62.51
        & {\textbf{45.81}}
        & 25.04
        & {\textbf{45.48}}
        & {\textbf{54.03}}\\
    
    \midrule
    
    \multirow{4}{*}{ResNet-101 + FPN}
    
        & Baseline
        & 42.28 & 62.65 & 46.03 & 25.32 & 45.77 & 54.42\\
    
        & Noise
        & 42.60 & 62.85 & 46.58 & 25.83 & 46.62 & 55.16\\
    
        & Union
        & 43.87 & \textbf{65.00} & 48.39 & \textbf{26.88} & 47.91 & 55.70\\
    
        & LossMix (ours)
        & \textbf{44.07} & 64.48 & \textbf{48.40} & 26.73 & \textbf{48.11} & \textbf{56.80}\\
    
    \bottomrule
    \end{tabular}
    \caption{
        MS COCO results with Faster RCNN  detector and ResNet-50/101 FPN backbone. Best checkpoints are selected according to the AP metric following MS COCO evaluation format. Models are trained for 270K iterations. Best numbers are in \textbf{bold}. The proposed LossMix outperforms the baseline and state-of-the-art methods for the majority of metrics.
    } 
    \label{tab:det_coco}
\end{table*}

\paragraph{MS COCO dataset}
Our results for MS COCO dataset is shown in \Tab{det_coco}
Here, we can see that the promising performance of LossMix on PASCAL VOC is also generalizable to a much bigger (10$\times$) dataset such as MS COCO as well.
In particular, our method again achieves the best overall AP scores at $41.82$ for ResNet-50 and $44.07$ for ResNet-101.
When considering all metrics, LossMix also outperforms the previous state-of-the-art mixing techniques in the majority of cases.
We believe these results, coupled with the simplicity of loss mixing operation, make LossMix an appealing alternative to the current unweighted union practice for data mixing in object detection.

\begingroup
\begin{table*}[t]
\centering
\begin{tabular}{@{}l|c|cccc|ccc@{}}
\toprule

~
        & \multicolumn{1}{c}{Input} 
        & \multicolumn{4}{c}{Loss mixing} 
        & \multicolumn{3}{c}{Evaluation} \\

Method
    & mixing
    & $\mathcal{L}_{cls}^{rpn}$
    & $\mathcal{L}_{reg}^{rpn}$
    & $\mathcal{L}_{cls}^{roi}$
    & $\mathcal{L}_{reg}^{roi}$
    & AP
    & AP$_{50}$
    & AP$_{75}$
    \\
    
\midrule\midrule

Baseline (no mixing)
    & \xmark
    & \xmark & \xmark & \xmark & \xmark 
    & 53.88 & 79.31 & 59.99 \\
Uniform Union (no loss mixing) 
    & $\lambda  \sim Beta(1.0,1.0) $
    & \xmark & \xmark & \xmark & \xmark 
    & 55.28 & 81.97 & 61.78 \\
    
\midrule
LossMix: ROI-only 
    & $\lambda  \sim Beta(1.0,1.0) $
    & \xmark & \xmark & \cmark & \cmark 
    & 56.15 & 82.07 & 62.90 \\
LossMix: Localization-only 
    & $\lambda  \sim Beta(1.0,1.0)$ 
    & \xmark & \cmark & \xmark & \cmark 
    & 55.31 & 82.15 & 62.61 \\
LossMix: Classification-only 
    & $\lambda  \sim Beta(1.0,1.0)$ 
    & \cmark & \xmark & \cmark & \xmark 
    & 56.55 & \textbf{82.71} & 63.38 \\

\midrule
LossMix: $\alpha=0.2$
    & $\lambda  \sim Beta(0.2,0.2) $
    & \cmark & \cmark & \cmark & \cmark 
    & 56.21 & 81.95 & 63.33 \\
LossMix: $\alpha=5.0$
    & $\lambda  \sim Beta(5.0,5.0) $
    & \cmark & \cmark & \cmark & \cmark  
    & 56.18 & 82.21 & 62.72 \\
LossMix: $\alpha=20.0$ 
    & $\lambda  \sim Beta(20.0,20.0)$
    & \cmark & \cmark & \cmark & \cmark 
    & 56.30 & 82.10 & 63.11 \\

\midrule
LossMix + RegMixup~\cite{regmixup}
    & $\lambda  \sim Beta(1.0,1.0) $
    & \cmark & \cmark & \cmark & \cmark 
    & 56.31 & 81.92 & 62.40 \\

LossMix + Early Stop~\cite{over_train_mixup})
    & $\lambda  \sim Beta(1.0,1.0) $
    & \cmark & \cmark & \cmark & \cmark 
    & 56.42 & 82.20 & 62.93 \\

\midrule
LossMix (default) 
    & $\lambda  \sim Beta(1.0,1.0)$ 
    & \cmark & \cmark & \cmark & \cmark 
    & \textbf{56.60} & 82.17 & \textbf{63.59} \\

\bottomrule
\end{tabular}

\caption{
    Ablation study on PASCAL VOC dataset. The base detector is Faster RCNN with ResNet-50 FPN backbone. Best AP checkpoints are reported. Best numbers are in \textbf{bold}. For early stopping, we train the model with LossMix for the first 16k iterations out of a total of 18k.
    The mixing coefficient $\lambda$ is sampled from $Beta(\alpha,\alpha)$ distribution, following Mixup.
} 
\label{tab:det_abl}
\end{table*}

\paragraph{Ablation study}
Although at its core, LossMix simply proposes the mixing of loss signals, there can be different implementation variations and hyper-parameters. 
\Tab{det_abl} provides an ablation study investigating how these options affect the performance of LossMix.
Overall, LossMix is robust with these configurations; all offer improvement over the Baseline (no data mixing) and the popular Union~\cite{union_da_fewshot_acrofod,union_instseg_seesaw,union_mot_bytetrack,bof,union_ssl_dual,union_ssl_instant_teaching} strategy.
Morevover, we can see that although mixing of classification losses ($\mathcal{L}_{cls}^{rpn}$ and $\mathcal{L}_{cls}^{roi}$) contributes the most, mixing box regression losses ($\mathcal{L}_{reg}^{rpn}$ and $\mathcal{L}_{reg}^{roi}$) can also help, yielding better localization results as shown by AP$_{75}$ as well as better overall AP.
It is important to highlight that even when incorporating only box classification losses, our proposed method goes beyond image-level Mixup.
This is because, by re-weighting $\mathcal{L}_{cls}$, LossMix effectively addresses a range of challenges related to spatial misalignment, background information, and object plurality that we have discussed in previous sections. 
In contrast, Mixup is not specifically designed to tackle these and cannot be adopted directly for object detection.
This underscores the distinct advantages of our approach.

\begingroup
\begin{table*}
\centering
\begin{tabular}{@{}l|cccc cccc |cc |c@{}}
\toprule
Method 
    & Source
    & SCL
    & SWDA
    & DM
    & CRDA
    & HTCN
    & UMT
    & AT*
    & Noise
    & Union
    & Ours\\
    
\midrule
\midrule
mAP
    & 28.8
    & 41.5
    & 38.1
    & 41.8
    & 38.3
    & 40.3
    & 44.1
    & 46.7
    & 44.9
    & 50.0
    & \textbf{51.1}\\

\bottomrule
\end{tabular}
\caption{
PASCAL VOC $\rightarrow$ Clipart1k adaptation results. The Average Precision (in \%) for all object classes from is reported, following \citet{umt} and \citet{amt}. The methods presented are SCL \cite{SCL}, SWDA \cite{SWDA}, DM \cite{DM}, CRDA \cite{CRDA}, HTCN \cite{HTCN}, UMT \cite{umt}, AT \cite{amt}, and Source (Faster-RCNN \cite{faster_rcnn}). Best results are in \textbf{bold}. *indicated reproduced results using the released code.
} 
\label{tab:voc_clip}
\end{table*}
\endgroup

\begingroup
\setlength{\tabcolsep}{5pt}
\begin{table}
\centering

\begin{tabular}{@{}l|cccccc|c@{}}
\toprule
Method 
    & bike & bird & car & cat & dog & person & mAP \\
\midrule
\midrule
Source 
    & 84.2 & 44.5 & 53.0 & 24.9 & 18.8 & 56.3 & 46.9 \\
\midrule
SCL 
    & 82.2 & 55.1 & 51.8 & \underline{39.6} & 38.4 & 64.0 & 55.2  \\
SWDA 
    & 82.3 & \textbf{55.9} & 46.5 & 32.7 & 35.5 & 66.7 & 53.3\\
DM 
    & - & - & - & - & - & - & 52.0 \\
UMT 
    & 88.2 & 55.3 & 51.7 & \textbf{39.8} & \textbf{43.6} & 69.9 & 58.1 \\
AT* 
    & \textbf{95.8} & 51.7 & \textbf{57.8} & 36.5 & 33.1 & 71.0 & 57.7 \\
\midrule
Ours
    & \underline{91.1} & \underline{55.8} & \underline{54.3} & 39.1 & \underline{41.0} & \textbf{74.3} & \textbf{59.3} \\

\bottomrule
\end{tabular}
\caption{
PASCAL VOC $\rightarrow$ Watercolor2k adaptation results. The Average Precision (in \%) is reported following \cite{amt}.
Best numbers are in \textbf{bold}.
2nd best are \underline{underlined}.
*indicated reproduced results using official code. 
} 
\label{tab:voc_water}
\end{table}
\endgroup

\section{Experiments: Domain Adaptation}
\subsection{Experimental Settings}
\paragraph{Datasets} We conduct our experiments for cross-domain object detection using two popular and challenging real-to-artistic adaptation setups~\cite{
HTCN,
umt,
DM,
amt,
SWDA,
SCL,
CRDA}:
PASCAL VOC~\cite{pascal} $\rightarrow$ Clipart1k~\cite{clipart_watercolor}
and
PASCAL VOC~\cite{pascal} $\rightarrow$ Watercolor2k~\cite{clipart_watercolor}.
Compared to PASCAL VOC, Clipart1k and Watercolor2k \cite{clipart_watercolor} represent large domain shifts from real-world photos to artistic images. 
Clipart1k dataset shares the same set of object categories as PASCAL VOC and contains a total of 1000 images. We split these into 500 training and 500 test examples. 
Watercolor2k dataset has 2000 images of objects from 6 classes in common with the PASCAL VOC. We also split the dataset in halves to obtain 1000 training and 1000 test images.

\paragraph{{Implementation Detail}} 
We leverage the open source code of the state-of-the-art Adaptive Teacher~\cite{amt} framework. 
The codebase is also built on top of Detectron2~\cite{detectron2}.
For fair comparison against previous works \cite{umt,amt}, we use Faster RCNN \cite{faster_rcnn} with ResNet-101 \cite{resnet} backbone. 
We follow the setup of \cite{amt} and scale all training images by resizing their shorter side to 600 while maintaining the image ratios. 
We keep all loss weight for labeled and pseudo-labeled examples to be 1.0 for simplicity and use the default weight of 0.1 for the discriminator branch. We also keep the confidence threshold as 0.8. 
We notice that the set of hyper-parameter reported in the \cite{amt} is not suitable for the open-sourced code. 
Thus, we tune them to get the best performance for Adaptive Teacher for fair comparison and keep their original set of strong-weak augmentations.

\subsection{Results}
\paragraph{PASCAL VOC $\rightarrow$ Clipart1k}
We compare with state-of-the-art methods in cross-domain object detection using the popular PASCAL VOC $\rightarrow$ Clipart1k adaptation (\Tab{voc_clip}).
We report an mAP of 50.33\% across all object categories, achieving the new state-of-the-art performance with +3.5\% improvement on top of the prior state of the art set by the recent Adaptive Teacher~\cite{amt}.
Despite AT's strong performance, our results suggest that large domain shifts are still challenging and reveal potential biases toward source domain, e.g. inherently in the warmup procedure of Mean Teacher.
By strategically leveraging LossMix, we are able to mitigate these problems and further improve accuracy.

\paragraph{Comparing with SOTA mixing}
\Tab{voc_clip} also presents our comparison to different Mixup variations used by existing methods, namely Union~\cite{union_da_fewshot_acrofod,union_instseg_seesaw,union_mot_bytetrack,bof,union_ssl_dual,union_ssl_instant_teaching} and Noise~\cite{afan}.
Specifically, AFAN \cite{afan} deploys a small $\lambda$ value on target domain image without any pseudo labels. This strategy is similar to our noise mixing during the warmup, but is used throughout the training.
Note that this approach performs worse than our AT basedline.
This is because although noise mixing could be helpful in general, as shown by both AFAN \cite{afan} and our following ablation studies, heavily relying on it in the adaptation phase of Mean Teacher can lead to bias towards the source domain due to the fact that ``mixed-in" target information is only limited to a tiny amount to act as a domain-aware augmentation. Indeed, we believe for cross-domain mean teacher, the pseudo labels are much stronger target signals and should be taken advantage of appropriately.
We also see sub-optimal results for Union~\cite{bof} due to errors in the approximation of unweighted union, similar to detection experiments.
\Tab{da_abl} shows an ablation study for more insights.

\begingroup
\setlength{\tabcolsep}{4pt}
\begin{table}
    \centering
    \resizebox{\linewidth}{!}{ 
    \begin{tabular}{@{}l|ccccccc|c@{}}
    \toprule
    ~
        & $\alpha$
        & warm
        & adapt
        & $\mathcal{L}_{cls}^{rpn}$
        & $\mathcal{L}_{reg}^{rpn}$
        & $\mathcal{L}_{cls}^{roi}$
        & $\mathcal{L}_{reg}^{roi}$
        & AP$_{50}$
        \\
    \midrule\midrule
    AT Baseline
        & ~ 
        & ~
        & ~
        & ~
        & ~
        & ~
        & ~
        & 46.7 \\
    \midrule
    LossMix: ROI 
        & 1.0
        & \cmark
        & \cmark 
        & ~
        & ~
        & \cmark 
        & \cmark
        & 49.6 \\
    LossMix: Loc 
        & 1.0
        & \cmark
        & \cmark
        &
        & \cmark
        &
        & \cmark
        & 48.2 \\
    LossMix: Cls 
        & 1.0
        & \cmark
        & \cmark
        & \cmark
        &
        & \cmark
        & 
        & 48.1 \\
    \midrule
    LossMix: warm 
        & 1.0
        & \cmark
        &
        & \cmark
        & \cmark
        & \cmark
        & \cmark
        & 49.1 \\
    LossMix: adapt 
        & 1.0
        &
        & \cmark
        & \cmark
        & \cmark
        & \cmark
        & \cmark
        & 48.1 \\
    \midrule
    LossMix: $\alpha$=0.2
        & 0.2
        & \cmark
        & \cmark
        & \cmark
        & \cmark
        & \cmark
        & \cmark
        & 47.8 \\
    LossMix: $\alpha$=5.0
        & 5.0
        & \cmark
        & \cmark
        & \cmark
        & \cmark
        & \cmark
        & \cmark
        & 49.8 \\
    LossMix: $\alpha$=20.0
        & 20.0
        & \cmark
        & \cmark
        & \cmark
        & \cmark
        & \cmark
        & \cmark
        & 49.4 \\
    \midrule
    LossMix (default) 
        & 1.0
        & \cmark
        & \cmark
        & \cmark
        & \cmark
        & \cmark
        & \cmark
        & \textbf{51.1} \\
    \bottomrule
    \end{tabular}
    } 
    \caption{
        Ablation study for PASCAL VOC $\rightarrow$ Clipart1k.
    } 
    \label{tab:da_abl}
\end{table}

\paragraph{PASCAL VOC $\rightarrow$ Watercolor2k}
Next, we are interested in answering the question whether or not the encouraging gains observed in PASCAL VOC $\rightarrow$ Clipart1k can be reproduced on a different dataset. To do this, we use Watercolor2k and evaluate the performance of PASCAL VOC $\rightarrow$ Watercolor2k adaptation.
Note that after experimenting with Clipart1k, we narrowed down our set of hyper-parameters to ones that work best for both Adaptive Teacher and our method for fair competition.
For Watercolor2k, to test our method's robustness, we directly perform grid search on this small set of hyper-parameters without any further tuning or manual supervision.
Nonetheless, even without exhaustive tuning, our results in \Table{voc_water} show that we can still outperform AT (mAP=57.7) and archive mAP=59.3 (+1.5 gain).

\section{Conclusion}
We tackle the challenges of applying data mixing augmentations to object detection.
Specifically, we introduce Supervision Interpolation (SI),
a novel conceptual reinterpretation and generalization of Mixup. 
Given SI, we propose LossMix, a simple-yet-effective regularization that interpolates the losses instead of labels to enhance model learning.
Our experiments show consistent accuracy improvements, outperforming popular object mixing strategies and achieving state-of-the-art domain adaptation results.
We hope this inspires future data mixing research for detection and beyond.

\bibliography{aaai24}

\clearpage
\appendix

\twocolumn[
\centering
\Large
\vspace{0.75em}
\textbf{Supervision Interpolation via LossMix:\\Generalizing Mixup for Object Detection and Beyond} \\
\vspace{0.75em}
{ Supplementary Material} \\
\vspace{1.5em}
] 

\appendix

\setcounter{page}{1}

In this section, we provide additional information on the following supplementary materials:

\begin{itemize}
    \item LossMix implementation
    \item Details of object detection experiments
    \item Domain adaptation training
    \item Details of domain adaptation experiments
    \item Detailed results on VOC $\rightarrow$ Clipart1k (\Table{voc_clip0})
    \item Qualitative Results
\end{itemize}

\begingroup
\setlength{\tabcolsep}{2pt}
\begin{table*}
    \centering
    \resizebox{\linewidth}{!}{ 
    \begin{tabular}{@{}l|cccccccccccccccccccc|c@{}}
    \toprule
    Method 
        & aero & bike & bird & boat & bottle & bus & car & cat & chair & cow & table & dog & hrs & mbike & prsn & plnt & sheep & sofa & train & tv & mAP \\
        
    \midrule
    \midrule
    Source
        &23.0 & 39.6 & 20.1 & 23.6 & 25.7 & 42.6 & 25.2 & 0.9 & 41.2 & 25.6 & 23.7 & 11.2 & 28.2 & 49.5 & 45.2 & 46.9 & 9.1 & 22.3 & 38.9 & 31.5 & 28.8 \\
        
    \midrule
    SCL
        & \textbf{44.7}& 50.0& 33.6& 27.4& 42.2& 55.6& 38.3& \textbf{19.2}& 37.9& \textbf{69.0}& 30.1& {26.3}& 34.4& 67.3& 61.0& 47.9& 21.4& 26.3& 50.1& 47.3& 41.5\\
    SWDA
        & 26.2& 48.5& 32.6& 33.7& 38.5& 54.3& 37.1& 18.6& 34.8& 58.3& 17.0& 12.5& 33.8& 65.5& 61.6& 52.0& 9.3& 24.9& {54.1}& 49.1& 38.1 \\
    DM
        & 25.8& 63.2& 24.5& 42.4& 47.9& 43.1& 37.5& 9.1& 47.0& 46.7& 26.8& 24.9& 48.1& {78.7}& 63.0& 45.0& 21.3& 36.1& 52.3& {53.4}& 41.8\\
    CRDA
        & 28.7& 55.3& 31.8& 26.0& 40.1& 63.6& 36.6& 9.4& 38.7& 49.3& 17.6& 14.1& 33.3& 74.3& 61.3& 46.3& 22.3& 24.3& 49.1& 44.3& 38.3\\
    HTCN
        & 33.6& 58.9& 34.0& 23.4& 45.6& 57.0& 39.8& 12.0& 39.7& 51.3& 21.1& 20.1& 39.1& 72.8& 63.0& 43.1& 19.3& 30.1& 50.2& 51.8& 40.3\\
    UMT
        & 39.6& 59.1& 32.4& 35.0& 45.1& 61.9& 48.4& 7.5& 46.0& 67.6& 21.4& \textbf{29.5}& 48.2& 75.9& 70.5& 56.7& {25.9}& 28.9& 39.4& 43.6& 44.1 \\
    AT
        &33.8 & 60.9 & 38.6 & \textbf{49.4} & 52.4 & 53.9 & \textbf{56.7} & 7.5 & 52.8 & 63.5 & 34.0 & 25.0 & \textbf{62.2} & 72.1 & 77.2 & 57.7 & \textbf{27.2} & \textbf{52.0} & \textbf{55.7} & 54.1 & 49.3 \\
    
    AT*
        &42.1 &69.1 &32.5 &{46.2} &52.4 &{71.5} &47.9 &13.1 &55.4 &44.2 &{34.5} &22.8 &58.0 &66.1 &66.4 &\textbf{63.2} &22.6 &29.2 &49.8 &48.5 &46.7\\
    
    \midrule
    
    Noise
        &40.6  &\textbf{70.5}  &35.1  &41.0  &47.5  &54.7  &49.8  &15.9  &61.4  &31.5  &41.2  &17.1  &51.4  &70.6  &61.6  &57.0  &21.3  &35.4  &46.5  &48.6  &44.9\\
    
    Union
        &44.1 &64.7 &44.8 &43.7 &60.3 &\textbf{72.1} &49.5 &13.3 &61.4 &53.0 &\textbf{50.9} &21.9 &59.0 &75.2 &74.7 &56.1 &17.3 &38.6 &51.3 &48.6 &50.0 \\
    
    \midrule
    
    Ours
        &44.4 &65.9 &\textbf{45.7} &42.7 &\textbf{63.2} &63.2 &{50.2} &13.3 &\textbf{62.7} &55.0 &29.3 &26.0 &{60.9} &\textbf{85.8} &\textbf{77.4} &62.0 &20.1 &{47.7} &51.0 &\textbf{55.3} &\textbf{51.1}\\
    
    \bottomrule
    \end{tabular}
    } 
    
    \caption{
    The experimental results of cross-domain object detection on the PASCAL VOC $\rightarrow$ Clipart1k adaptation. The average precision (AP, in \%) for all object classes from is reported, following \cite{amt,umt}. 
    The methods presented are SCL \cite{SCL}, SWDA \cite{SWDA}, DM \cite{DM}, CRDA \cite{CRDA}, HTCN \cite{HTCN}, UMT \cite{umt}, AT \cite{amt}, and Source (Faster-RCNN \cite{faster_rcnn}).
    Best results are in \textbf{bold}.
    *indicated reproduced results using the authors' released code.
    Although we include the original numbers cited in AT paper~\cite{amt} for completeness,
    the author acknowledged there are existing instability issues with the open-sourced repository and could only acquire up to AP=45.6\% themselves. 
    } 
    \label{tab:voc_clip0}
\end{table*}
\endgroup

\section{LossMix Implementation}
\paragraph{Data Selection} In our implementation of LossMix, we follow the recommendation of Mixup~\cite{mixup} and mix two data points/images together. We did not experiment with mixing three or more data points/images. To minimize the I/O requirements, we use a single data loader. Each minibatch is then mixed with a randomly shuffled version of itself. 
Another implementation that uses a single loader is to reverse the minibatch, as used by {FixMatch}~\cite{fixmatch}, or to shift all elements by one index.

\paragraph{Input Mixing}
For each pair of data points, we use a $\text{Beta}(\alpha, \alpha)$ distribution to sample a mixing coefficient $\lambda$. 
We set $\alpha$ to $1.0$ by default,
following \cite{bof}.
This results in a uniform sampling distribution for $\lambda$.
When mixing two images, we first create a mixed output image with dimensions equal to the maximum height and width of the original images, then apply random translation, within the output canvas, for each input image.
We pad the resulting image with zeros to maintain the original aspect ratios, which is important for preserving the aspect ratios and geometry of instances in object detection.
We do not experiment with other alignment options, such as aligning by the image origin $(0, 0)$ or using random translations. The effects of different $\alpha$ values on LossMix are presented in \Table{det_abl}.

\paragraph{Target Mixing}
According to \Equation{our_y}, the augmented ground truth for the mixed image is an implicitly weighted union of the original ground truth: $\tilde{y} = \{(y_i; \lambda), (y_j; (1 - \lambda)\}$, where $\lambda$ is the mixing coefficient sampled from a Beta distribution as described in the previous paragraph.
In the case of object detection, the label $y$ is a set of instance ground truth, which is a collection of objects in the image, each with its corresponding class labels and bounding box coordinates.
For example, $y_i$ could be a set of objects $\{(c_{i1},b_{i1}), (c_{i2},b_{i2}),\ldots\}$ present in the image $x_i$.
$c_{ik}$ denotes the class label of object $k$ and $b_{ik}$ represents its corresponding bounding box coordinates.
After mixing, all objects of $y_i$ will share the same mixing weight $\lambda$ of the image, while those of $y_j$ will have weight $(1 - \lambda)$.
These instance-level weights will then be used for loss mixing as described in \Equation{our_loss} and \Equation{det_loss_mix}.
Finally, all bounding box coordinates are adjusted appropriately based on the alignment operation in the \textit{Input Mixing} step described above.

\section{Object Detection Experiments}
We leverage the open-source PyTorch-based Detectron2~\cite{detectron2} as our codebase for object detection experimentation.
Faster RCNN~\cite{faster_rcnn} with ResNet~\cite{resnet}--FPN~\cite{fpn} backbone is employed as our baseline architecture.
All backbones use Synchronized Batch Normalization, which we found to be either on par or better then frozen BatchNorm in most cases.
Unless otherwise specified, we follow most of the default configurations of Detectron2 for both PASCAL VOC~\cite{pascal} and MS COCO~\cite{coco} datasets. 
Specifically, we use ImageNet1K~\cite{imagenet} pretrained weights to initialize the ResNet-50 and ResNet-101~\cite{resnet} backbones.
We use Stochastic Gradient Descent (SGD) optimizer
and random horizontal flipping.
We use a batch size of 64 for faster convergence, an initial learning rate of 0.08, and the default step scheduler from Detectron2.
We use linear warm-up for $100$ iterations with a warm-up factor of $0.001$.
All models use multi-scale training with smallest image side randomly sample from $(480,\ldots, 800)$ for VOC and $(640,\ldots, 800)$ for COCO, both with an increment of $32$. 
The minimum image side at test time is set to $800$ and the maximum for both training and testing is $1333$ by default.
We train PASCAL VOC for 18,000 iterations, which is about 70 epochs, and MS COCO for 270,000 iterations ($3\times$ schedule), equating about 146 epochs.
All experiments were trained with 8 NVIDIA GPUs, either V100 or A100.

Regarding the implementation of other methods, we sample $\lambda \sim \text{Beta}(1.0, 1.0)$ for the popular unweighted \textit{Union} mixing strategy~\cite{union_da_fewshot_acrofod,union_instseg_seesaw,union_mot_bytetrack,bof,union_ssl_dual,union_ssl_instant_teaching}, similar to the setting for our LossMix.
For \textit{Noise} mixing strategy~\cite{afan}, we randomly sampled $\lambda \sim \text{U}(0.0, 0.2)$, i.e., a small mixing ratio from an uniform distribution with an arbitrary upper bound of $0.2$ and only keep the instance labels of the image with the larger coefficient $(1-\lambda)$.
For LossMix-Reg model in \Table{det_abl} deploying LossMix in a RegMixup~\cite{regmixup}-style, we set half of the total batch size to be mixed data and the other half to be regular, non-mixed data, which have effective loss weights of $\lambda=1.0$.

\begin{figure*}
    \begin{center}
    \includegraphics[width=0.9\linewidth]{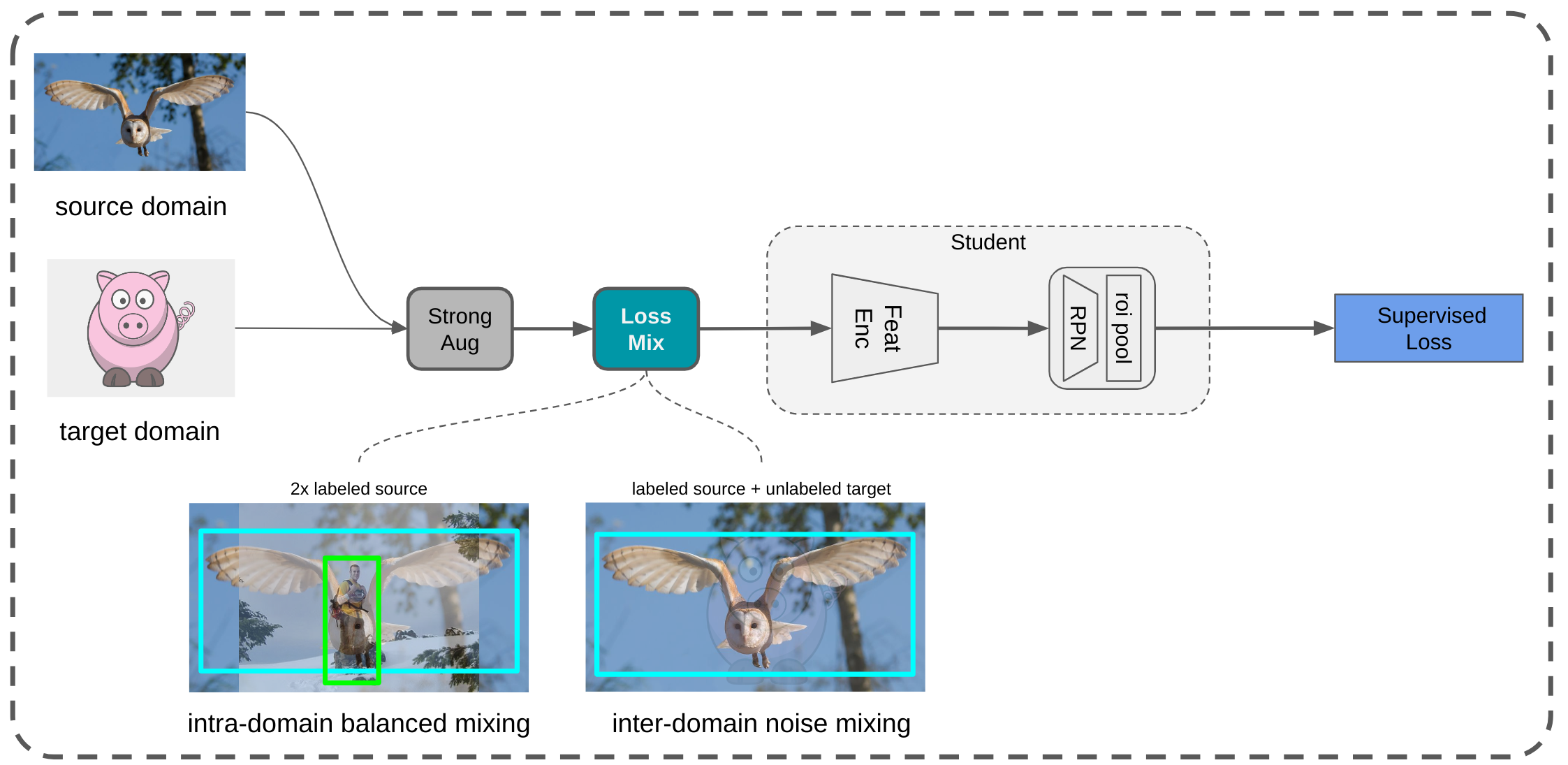}
    \\
    {Warmup Phase}
    \\
    \vspace{0.75em}
    \includegraphics[width=0.9\linewidth]{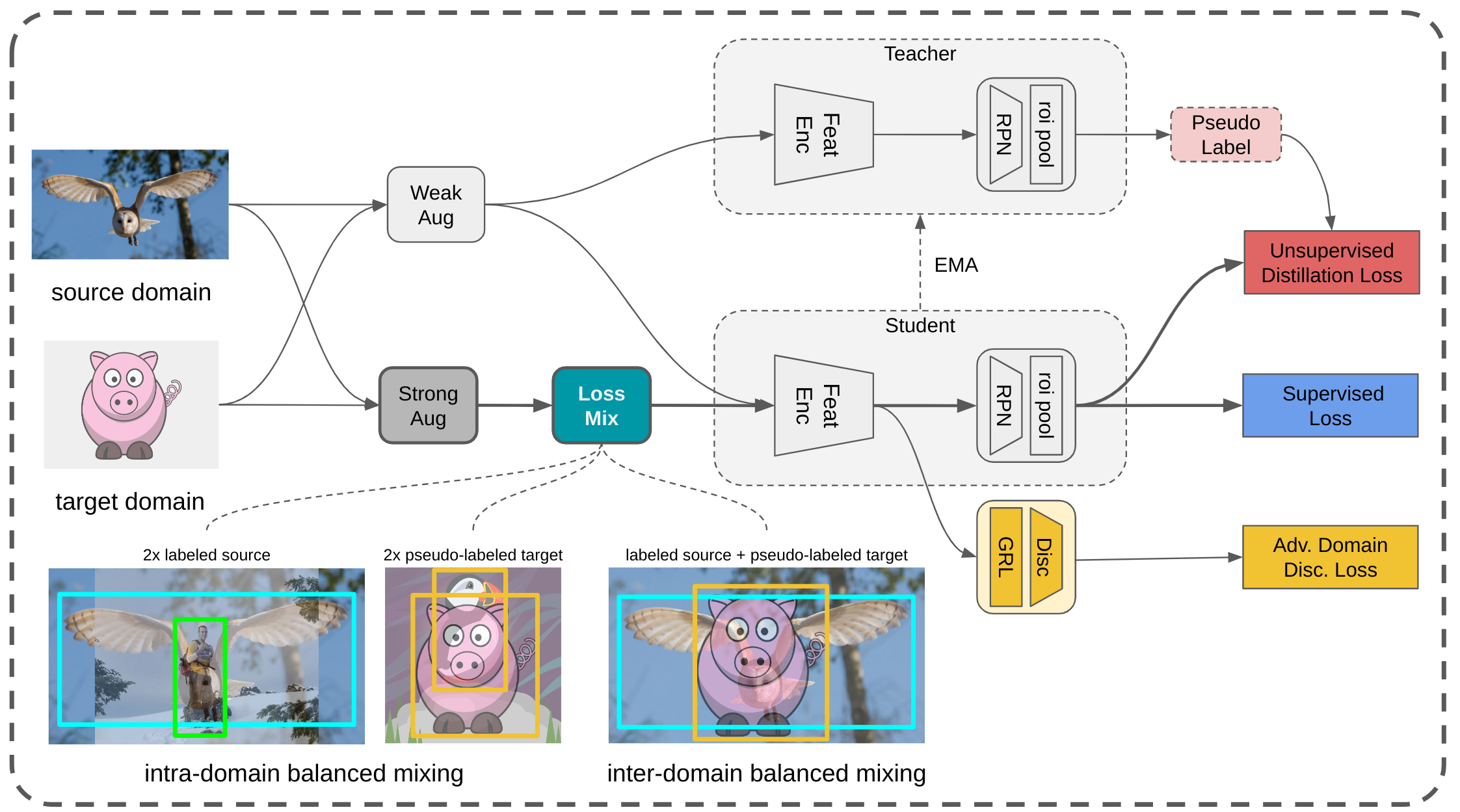}
    \\
    {Adaptation Phase}
    \\
    \vspace{0.75em}
    \includegraphics[width=0.9\linewidth]{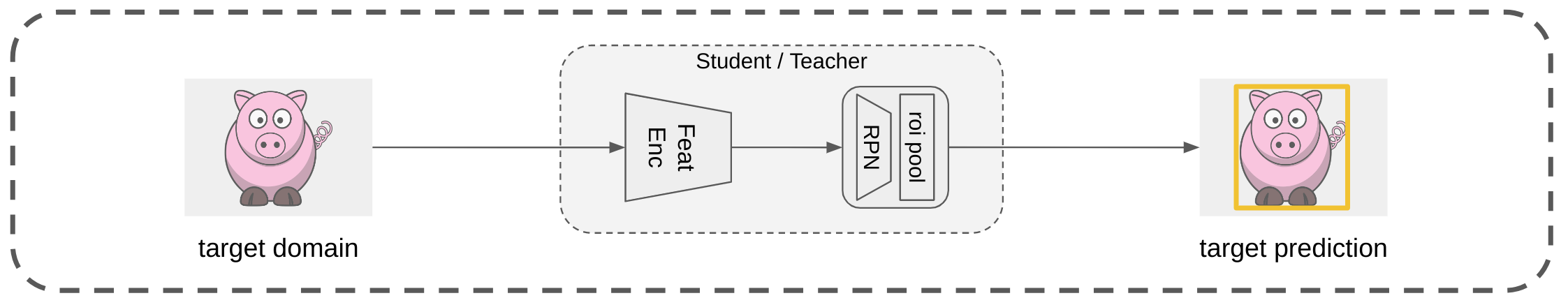}
    \\
    {Inference Time}
    \end{center}
    \caption{
    Overview of our mixed-domain teacher model during three different phases: Warmup, Adaptation, and Inference.
    }
    \label{fig:phases}
\end{figure*}

\section{Domain Adaptation Training}
\Figure{phases} provides a visualization of the Mean Teacher training process using our LossMix approach and different domain mixing strategies.
During the Warmup phase, only the Student network is trained using a supervised loss, such as cross-entropy. In this phase, we do not use pseudo-labels generated by the Teacher network or adversarial learning.
We employ two different mixing strategies: (a) intra-domain source$\times$source mixing with ground truth annotations, and (b) inter-domain noise mixing with labeled source data and unlabeled target images.
The warmup loss function can be defined as follows:
\begin{equation} \label{eq:loss_warm}
    \mathcal{L}_{warm} = \lambda_{mss}\mathcal{L}_{mss} + \lambda_{nst}\mathcal{L}_{nst}
\end{equation}
where $\mathcal{L}$ is short for the detection loss $\mathcal{L}_{det}$ in \Equation{det_loss_mix}.
the term ``mss" denotes mixed source$\times$source data, where two labeled source images are randomly selected and mixed together. Similarly, ``nst" refers to noise-mixed source$\times$target data, where a labeled source image and an unlabeled target image are mixed together.

During the Adaptation phase, we utilize three mixing procedures to further enhance the training of the Mean Teacher model: (a) intra-domain labeled source$\times$source mixing, (b) intra-domain pseudo-labeled target$\times$target mixing, and (c) inter-domain source$\times$target mixing. 
Importantly, all mixing operations are performed in a balanced manner, such that both source labels and target pseudo-labels are treated equally during mixing. Furthermore, no noise mixing is used during this phase. 
The adaptation loss function can be written as follows:
\begin{align} \label{eq:loss_adapt}
\begin{split}
    \mathcal{L}_{adapt} &= \lambda_{mss}\mathcal{L}_{mss} + \lambda_{mtt}\mathcal{L}_{mtt} \\
    &+ \lambda_{mst}\mathcal{L}_{mst} + \lambda_{disc}\mathcal{L}_{disc}
\end{split}
\end{align}
The term ``mtt" denotes mixed target$\times$target data, ``mst" denotes balanced (as opposed to noise mixing) mixed source$\times$target data, and $\mathcal{L}_{disc}$ refers to the adversarial domain discriminator loss~\cite{amt}.
The pseudo-labeling losses for ``mtt" and ``mst" is implemented with cross-entropy loss function, as in prior studies \cite{umt,amt}.

Finally, no augmentation is used during inference so the model's speed is not negatively affected in any capacity when training with our proposed method. Moreover, one advantage of Mean Teacher~\cite{umt,amt,mean_teacher} framework, which models trained with our LossMix also inherit, is that the Teacher model shares the exact same architecture as its Student. The only difference is their learned weights. Therefore, after training, we have the option to safely select the model weights with higher performance among the two for deployment, without any additional latency cost.

\section{Domain Adaptation Experiments}
For our domain adaptation experiments, we use the official open-source repository published by Adaptive Teacher~\cite{amt}, which is also built with Detectron2~\cite{detectron2}.
For a fair comparison, we follow previous works~\cite{umt,amt} and use Faster RCNN~\cite{faster_rcnn} with a ResNet-101~\cite{resnet} backbone. 
All images are resized to have a shorter side of $600$ while maintaining their aspect ratios. The confidence threshold for filtering pseudo labels is set to $0.8$. 
Weak augmentations include random horizontal flipping and cropping, while strong augmentations include random color jittering, grayscaling, Gaussian blurring, and cutout. The weight smoothing coefficient of the exponential moving average (EMA) for updating the Teacher model is set to $0.9996$. For simplicity, we keep all loss weights for labeled and pseudo-labeled examples (\Equation{loss_warm} and \Equation{loss_adapt}) at $1.0$ and do not tune them. We use Adaptive Teacher's default weight of $0.1$ for the discriminator branch.

We found that the set of training hyperparameters reported in Adaptive Teacher \cite{amt} yield suboptimal results for the open-sourced codebase.
The authors confirmed that there are instability issues\footnote{https://github.com/facebookresearch/adaptive\_teacher/ issues/26\#issuecomment-1192059882} 
and were only able to achieve an mAP of $45.6$\footnote{https://github.com/facebookresearch/adaptive\_teacher/ issues/9\#issuecomment-1193174238} 
instead of the reported mAP of $49.3$ in their paper \cite{amt}. 
This is because the open-sourced code is built with Detectron2 \cite{detectron2}, while their original internal code was built with D2GO\footnote{https://github.com/facebookresearch/adaptive\_teacher/ issues/9\#issuecomment-1134933287}.
To ensure a fair comparison, we tuned the hyperparameters and report our best mAP number of $46.7$ for Adaptive Teacher in Table \Table{voc_clip}. 
Specifically, we performed a grid search over batch sizes of $\{8,2\}$ and learning rates of $\{0.002, 0.005, 0.01\}$ for both the baseline Adaptive Teacher and our model trained with LossMix. The total training consisted of 60,000 iterations, with 20,000 iterations for warm-up and 40,000 iterations for adaptation. We observed that most models reach their peak performance within the first 20K steps of the adaptation phase (after the warm-up phase). 
If the instability issues are resolved, the results for both the reproduced AT (mAP=$46.7$) and our LossMix (mAP=$51.1$) in  \Table{voc_clip} could potentially improve.

\section{Qualitative Results}
\Figure{qualitative_clipart} illustrates the qualitative results of the PASCAL VOC~\cite{pascal}~$\rightarrow$~Clipart1k~\cite{clipart_watercolor} adaptation. To facilitate comparison, we also display the ground truth labels and predictions of Adaptive Teacher~\cite{amt}. The threshold for category scores is set to 0.6 to enhance visualization.

\begin{figure*}
    \begin{center}
    \includegraphics[width=0.77\linewidth]{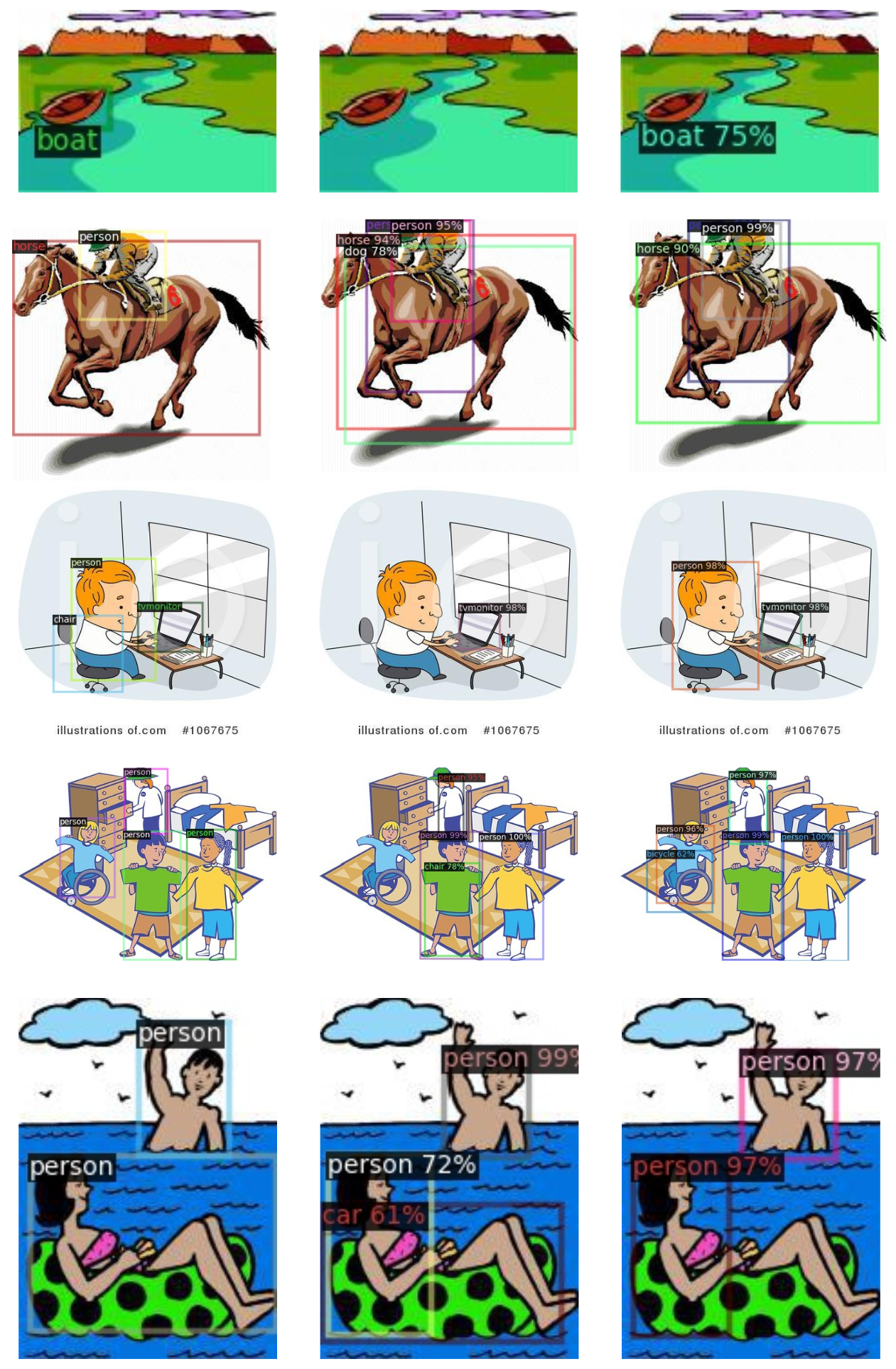}
    \end{center}
    {
        \hspace{9.3em}
        Ground truth 
        \hspace{6.6em}
        Adaptive Teacher
        \hspace{8.3em}
        Ours
    }
    \vspace{1em}
    \caption{
    Qualitative results for the PASCAL VOC~\cite{pascal}~$\rightarrow$~Clipart1k~\cite{clipart_watercolor} adaptation.
    } 
    \label{fig:qualitative_clipart}
\end{figure*}

\end{document}